\documentclass[]{article}
\usepackage{cite}
\usepackage{amsmath,amssymb,amsfonts}
\usepackage{algorithmic}
\usepackage{graphicx}
\usepackage{multirow}
\usepackage{array}
\usepackage{textcomp}
\usepackage{subfigure}
\usepackage{bm}
\usepackage{fancyhdr} 
\usepackage{hyperref} 
\usepackage{eso-pic}
\usepackage{lipsum} 

\title{Dynamic Classification: Leveraging Self- supervised Classification to Enhance Prediction Performance}
\author{First A. ZIYUAN ZHONG, zzy13534959496@163.com \\ 
	Second B. JUNYANG ZHOU, junyang.zhou@gmail.com}

\AddToShipoutPicture{%
	\AtPageLowerLeft{%
		\raisebox{\dimexpr\paperheight-1cm}{
			\makebox[2cm][l]{
			\underline{	This ariticle has been accepted by IEEE ACCESS, DOI:10.1109/ACCESS.2025.3575232}}
		}%
	}%
}

\begin{document}
\maketitle

\newpage
\begin{abstract}
In this study, we propose an innovative dynamic classification algorithm aimed at achieving zero missed detections and minimal false positives, critical in safety-critical domains (e.g., medical diagnostics) where undetected cases risk severe outcomes. The algorithm partitions data in a self-supervised learning-generated way, which allows the model to learn from the training set to understand the data distribution and thereby divides training set and test set into N different subareas. The training and test subsets in the same subarea will have nearly the same boundary. For each subarea, there will be the same type of model, such as linear or random forest model, to predict the results of that subareas. In addition, the algorithm uses subareas boundary to refine predictions results and filter out substandard results without requiring additional models. This approach allows each model to operate within a smaller data range and remove the inaccurate prediction results, thereby improving overall accuracy. Experimental results show that, with minimal data partitioning errors, the algorithm achieves exceptional performance with zero missed detections and minimal false positives, outperforming existing ensembles like XGBoost or LGBM model. Even with larger classification errors, it remains comparable to that of state-of-the-art models.

Key innovations include self-supervised classification learning, small-range subset predictions, and optimizing the prediction results and eliminate the unqualified ones without the need for additional model support. Although the algorithm still has room for improvement in automatic parameter tuning and efficiency, it demonstrates outstanding performance across multiple datasets. Future work will focus on optimizing the classification components to enhance robustness and adaptability.

\end{abstract}


\newpage
\section{Introduction}
The rapid increase in global energy demand has not only exacerbated environmental pollution but also intensified climate change and deepened the energy crisis. Consequently, the development of renewable energy sources has become a crucial direction for achieving sustainable global development \cite{miao2020}. Currently, renewable energy sources such as solar, wind, and biomass are gradually becoming key alternatives to fossil fuels. However, as of 2017, fossil energy still accounted for 73.5\% of the global electricity supply, whereas renewable energy accounted for only 26.5\% \cite{qazi2019}. Promoting the research, development, and application of new energy technologies has been recognized as a core strategy for addressing future energy shortages and environmental challenges.

Energy storage technology, particularly battery technology, plays a critical role in new energy systems. It not only promotes the efficient use of renewable energy but also plays a vital role in transforming energy structures \cite{miao2020}. Lithium-ion batteries, liquid current batteries, sodium-sulfur batteries, and other energy storage technologies have been rapidly developed in recent years, showing significant improvements in energy density, safety, and cost. However, there remains a gap in realizing the goal: high safety, low cost, long lifespan, and environmental friendliness \cite{johansson2024}. Therefore, future technological developments in energy storage are expected to facilitate the coexistence of various approaches.

In recent years, with an in-depth study of battery materials, new battery technologies have emerged. Research has shown that optimizing the material composition and battery design can significantly enhance battery performance by increasing the energy density and cycle life \cite{miao2020}. Furthermore, the introduction of the circular economy concept has provided new directions for the sustainable development of batteries, including efficient material recycling and reuse \cite{johansson2024}. Nonetheless, battery technology still faces many challenges in achieving the goals of a fully sustainable lifecycle and environmental protection. Future research should focus on discovering new materials and the synergistic application of multiple technological approaches \cite{miao2020, johansson2024}.

Artificial intelligence (AI) has significant potential in the field of energy storage. AI has accelerated the discovery, performance prediction, and optimal design of new materials using machine learning and big data technologies. For instance, a material genome program developed using AI has significantly improved the efficiency of battery materials development \cite{luo2020}. In addition, AI has demonstrated strong capabilities in battery condition monitoring, fault diagnosis, and performance prediction, providing robust support for advancing battery technology \cite{luo2020}. In the future, AI is expected to drive innovation and widespread adoption of battery technologies by, injecting new vitality into energy storage development \cite{luo2020}.

The rapid advancement of AI technology has offered new opportunities for innovation in battery technology. In recent years, machine learning algorithms have been widely applied in battery performance prediction and state-of-health (SOH) estimation, primarily focusing on noninvasive methods to obtain microstate information inside batteries \cite{jones2022}. For example, using data-driven machine learning models, the charging and discharging curve characteristics can be analyzed to predict the capacity degradation and remaining battery lifetime \cite{durmus2024}. Moreover, AI has accelerated the development of novel energy storage materials through high-throughput computing and material genome programs, thereby significantly improving research efficiency and material screening accuracy \cite{luo2020}.

Regarding application of AI in lithium-ion batteries, current research mainly focuses on model-based and data-based approaches \cite{durmus2024}. Model-based methods are used to analyze the chemical reactions and degradation mechanisms of the batteries. Although theoretically reflective of battery operations, these models often suffer from large errors due to complex internal chemical reactions and environmental factors \cite{jones2022}. Data-based methods, such as convolutional neural networks (CNN), recurrent neural networks (RNN), and support vector machines (SVM), have attracted attention owing to their flexibility and accuracy in nonlinear problems \cite{durmus2024}. For instance, CNN and RNN models optimized by genetic algorithms (GA) have significantly improved battery capacity prediction, achieving a root mean square error (RMSE) of only 0.1176\%, representing a considerable advancement over the previous models.

Despite these advances, several challenges remain. The predictive performance of data-driven approaches relies heavily on high-quality data, which are costly and limited in quantity\cite{jones2022}. In addition, owing to the different internal structures and operating conditions of different batteries, the generalization ability of these models is weak, making it difficult to adapt to complex and changing real-world situations\cite{miao2020}. Traditional methods usually rely on trial and error in parameter setting, which is a time-consuming and labor-intensive process that limits the practical application of these models\cite{durmus2024}. Finally, there is a need in the industry for a reliable method of picking out substandard products, and not just a need to make predictions about a particular batch. In other words, there is an industrial need for zero misses and fewer false positives. The current practice is to overlay the prediction with a classification model to pick out the substandard products, an approach that makes it difficult to strike a balance between misses and false kills, and is also unreliable.

In safety-critical domains, the stringent requirement for zero missed detection is intrinsically linked to system reliability and societal risk mitigation.In safety-critical domains, the stringent requirement for zero missed detection is intrinsically linked to system reliability and societal risk mitigation. Recent advances in self-supervised learning have demonstrated their potential in addressing distributional biases in industrial data. For instance, Zhang et al. \cite{zhang2024} proposed a debiased contrastive learning framework that effectively reduces sampling bias in time-series fault detection, achieving an F1-score of 97.2\%. Meanwhile, Wang et al. \cite{wang2025} developed a cognitive learning mechanism to dynamically adjust decision boundaries in multi-objective scenarios, showing a 7.58\% improvement in task-specific accuracy. Building on these advancements, our dynamic classification algorithm integrates self-supervised data partitioning with adaptive boundary refinement to simultaneously achieve zero missed detections and low false positives across complex industrial scenarios. In autonomous driving, Chen et al.  \cite{chen2022} demonstrated that perception models failing to detect pedestrians (with a missed detection rate exceeding \(10^{-6}\)) violate the ISO 21448 functional safety standards, potentially leading to fatal accidents. This finding echoes the foundational argument by Amodei et al. Similarly, in industrial manufacturing, Ren et al. \cite{ren2021} achieved zero-defect detection via deep learning, proving that a single missed defect could result in the scrapping of entire semiconductor batches, incurring multimillion-dollar economic losses. Collectively, these studies underscore that zero missed detection is not an aspirational goal but a non-negotiable constraint for safety-critical system design. Current technologies urgently require solutions such as the dynamic classification algorithm proposed in this work, which leverages self-supervised data partitioning and sub-domain model collaboration to simultaneously achieve zero missed detections and low false positives across complex scenarios---addressing cross-domain demands from autonomous driving to precision manufacturing.  

To address these challenges, this paper proposes a dynamic classification algorithm (DCA). The algorithm automatically divides the data into subsets using self-supervised learning and enables each subset to predict a small range of data based on the distributional features of the training set. Finally, the information from previous self-supervised learning is skillfully used to filter the predictions, which in turn reliably achieves the conditions of 0 misses and low false positives, thus eliminating the need for additional models. This approach significantly improves the performance when the classification error is small and rivals the current state-of-the-art models when the classification error is large. Dynamic classification algorithms are highly adaptable and expected to be widely used in various fields and industries.

\section{Related Work}

Numerous studies have explored methods to improve the model prediction accuracy, with a focus on optimization algorithms, machine learning, and hybrid approaches. This section reviews the key advancements in the field, highlighting their methodologies, strengths, and limitations, and provides a comparative analysis with the proposed Dynamic Classification Algorithm (DCA).

Azevedo et al. \cite{azevedo2023} conducted a systematic review of hybrid optimization and machine learning methods for clustering and classification. Their approach integrated various techniques to overcome the limitations of single methods. However, their framework requires significant computational resources and faces challenges when in adapting to rapidly evolving data environments. Kotary et al. \cite{kotary2023} proposed a joint prediction and optimization learning framework that directly learns optimal solutions from observable features. Despite its theoretical innovation, this approach demands end-to-end training for optimization problems, which can be computationally impractical, especially for nonconvex or discrete scenarios.

Khan et al. \cite{khan2023} combined neural networks with traditional filters (e.g., Kalman and $\alpha$-$\beta$ filters) to enhance the prediction accuracy of dynamic systems under noisy conditions. Although this method demonstrated significant improvements in dealing with uncertainty, it relied on complex parameter tuning, limiting its applicability in resource-constrained environments. Ippolito et al. \cite{ippolito2023} combined supervised and unsupervised learning to improve the stratigraphic classification accuracy, offering probability distributions rather than discrete classifications. However, their method requires extensive data preprocessing and feature engineering, which increases computational costs.

Son et al. \cite{son2023} applied both supervised and unsupervised learning to predict the social media user engagement in the airline industry. Although innovative in its use of sentiment analysis for classification, the model requires intensive data cleaning and customization in different business contexts. Shah and Satyanarayana \cite{shah2023} introduced a predictive range corrector model to enhance the reliability by accounting for deviations between the observed and predicted outputs. Although this approach improved prediction accuracy, its reliance on control sequence optimization made it less scalable for broader applications.

Dalal et al. \cite{dalal2023} developed a supply chain optimization model using Convolutional Neural Networks (CNNs) and Bidirectional Long Short-Term Memory Networks (BiLSTMs). Their method improves sustainability by leveraging CNNs for resource allocation and BiLSTMs for temporal demand forecasting. Similarly, Guillén et al. \cite{guillen2023} introduced the LSB-MAFS technique, combining Least Squares Boosting (LSB) and Multivariate Adaptive Regression Spline (MARS) to enhance production frontier forecasting. Although effective, the method primarily addressed specific regression tasks, limiting its adaptability.

Currently, the data prediction results are often filtered using additional classification models. However, in practical applications, these models typically struggle to establish a reliable balance between missed detections and false positives. This approach not only increases the computational overhead but also risks over-filtering or omitting crucial data, thereby compromising the overall reliability of the prediction system. In contrast, the proposed DCA employs self-supervised learning to automatically partition data subsets and, by predicting within small classification errors, directly filters results based on the distributional characteristics of the training set—without the need for additional models. The dynamic classification method ensures that the range for each subset is accurately defined, which allows the algorithm to achieve zero missed detections while effectively minimizing false positives, thereby improving the overall prediction accuracy. Furthermore, experimental results show that when the classification errors are minimal in the self-supervised subset partitioning phase, DCA significantly outperforms existing methods in terms of prediction stability and accuracy, and demonstrates strong adaptability across various datasets and application scenarios.
\newpage
\section{Abbreviations and Keywords List}

\begin{table}[h!]
	\centering
	\caption{Abbreviations and Keywords List}
	\label{tab:Abbreviations}
	\begin{tabular}{|p{80pt}|p{240pt}|}
		\hline
		\textbf{Abbreviation} & \textbf{Definition} \\ 
		\hline
		DC Error & Error rates during Dynamic Classification Process inside DC algorithm \\ 
		DC-E & Dynamic classification algorithm with result excluded \\ 
		Excluded Rate & Percentage of excluded data in the total volume \\ 
		$N$ & Number of classifications \\ 
		$Train_p$ & Data other than $Train_t$ in the training set \\ 
		$Train_t$ & Training-training set, the dataset obtained by re-sampling from the training set \\ 
		\hline
		$BP$ & Back Propagation Neural Network \\ 
		$DC$ & Dynamic classification algorithm \\ 
		$GC$ & Gaussian classification \\ 
		$GMM$ & Gaussian Mixture Model \\ 
		$KC$ & Kmeans classification \\ 
		$LGBM$ & Light Gradient Boosting Machine Model \\ 
		$N$ & Number of classifications \\ 
		$RF$ & Random Forest Model \\ 
		$SVM$ & Support Vector Machine Model \\ 
		$XGBoost$ & eXtreme Gradient Boosting \\ 
		\hline
	\end{tabular}
\end{table}

\section{Algorithm Structure Overview}
The DCA is designed to enhance prediction accuracy by partitioning the data into smaller, well-defined subsets. The algorithm consists of four core components, each of which contributes significantly to overall effectiveness:

\begin{itemize}
	\item \textbf{Routine Operations} – Standard preprocessing tasks such as data cleaning, normalization, and feature selection, which are essential in machine learning workflows.
	\item \textbf{Dynamic Classification Process} – The core innovation of DCA is that self-supervised learning iteratively refines data segmentation and improves classification boundaries.
	\item \textbf{Redundant Training and Prediction} – A mechanism that enhances stability by incorporating data from neighboring subsets, improves prediction accuracy.
	\item \textbf{Excluding Predicted Results} – A filtering strategy that removes predictions failing to meet accuracy requirements, ensuring that retained results achieve balanced zero missed detections and minimal false positives.
\end{itemize}

\begin{figure}[htbp]
	\centering
	\includegraphics[width=1\textwidth]{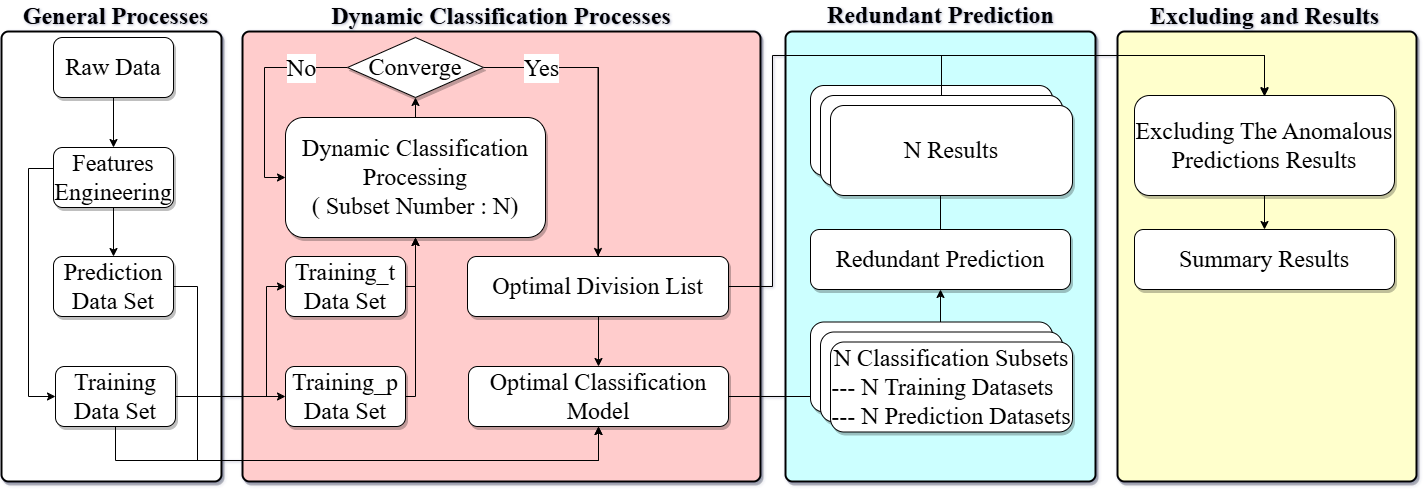} 
	\caption{\textbf{Structure of Dynamic Classification Algorithm}}
	\label{fig:figure1}
\end{figure}

This section provides a detailed breakdown of the algorithm structure, classification principles, redundant training mechanism, and filtering of the predicted results. The overall structure is illustrated in \textbf{Fig.\ref{fig:figure1}}, which divides the framework into four main components.

The **routine processing module (white section)** consists of fundamental preprocessing tasks such as data cleaning, normalization, and feature selection. These steps ensure data quality before classification and prediction.

The **dynamic classification module (red section)** is at the core of DCA. The dataset is split into $Train_t$ and $Train_p$. Through iterative optimization, the algorithm adjusts classification boundaries based on the numerical distribution of training data, ensuring that each subset operates within a smaller, controlled data range instead of relying on broad, generalized classification. The boundary information of these subsets is used as pseudo-labels for subsequent processing.

The **redundant training and prediction module (blue section)** further optimizes classification by incorporating data from adjacent classification intervals. This module enhances model robustness by supplementing predictions with contextual data and reducing boundary inconsistencies.

The **excluding predicted results module (yellow section)** uses  self-supervised learning pseudo-labels to filter out unreliable predictions. By leveraging pre-learned segmentation boundaries, this module ensures that only high-confidence predictions are retained, effectively achieving well-balanced zero missed detections and minimal false positives.

These four components collectively optimizing classification, improve prediction accuracy, and enhance model adaptability across various datasets. Given that the "routine processing module" section is centered on fulfilling routine tasks and does not incorporate any novel content, it will be omitted during the subsequent introduction of the principle of the DCA.

\subsection{Principle of Dynamic Classification Processes}
The **Dynamic Classification Process** is a fundamental component of the DCA. In traditional classification models, the learning process typically begins with extracting feature representations, followed by clustering or classification. However, this approach can lead to the coexistence of high and low values within the same classification, which diminishes the predictive model's ability to generalize effectively. Specifically, when target values span a wide numerical range, this inconsistency complicates thelearning process of the model, as the features may be similar while the predicted values diverge significantly.

\begin{figure}[htbp]
	\centering
	\includegraphics[width=1\textwidth]{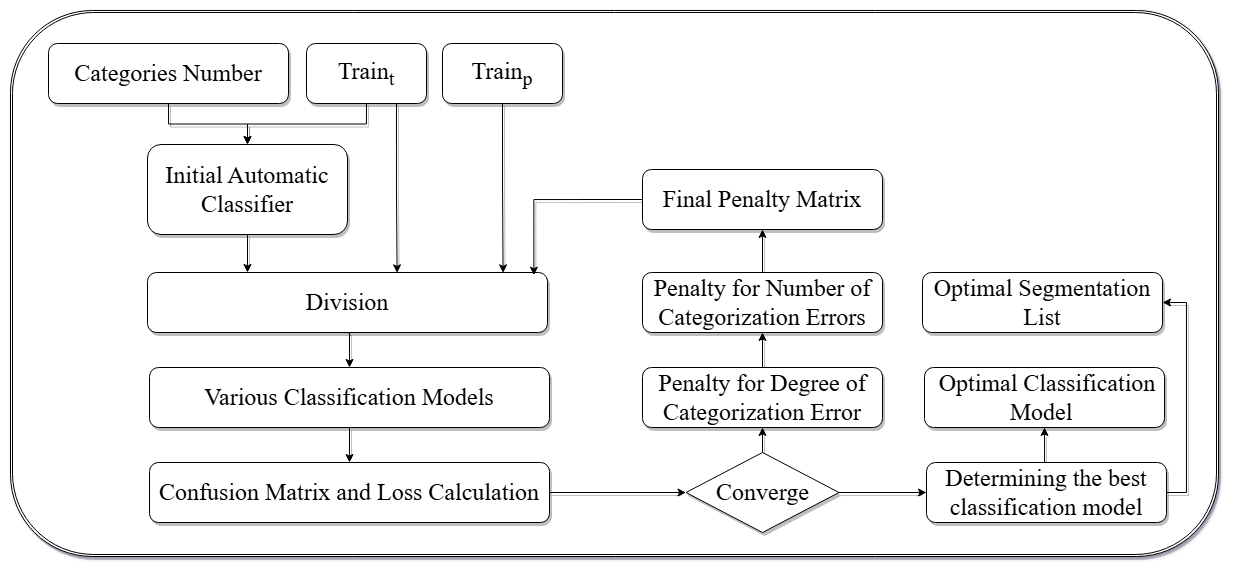} 
	\caption{\textbf{Structure of Dynamic Classification Algorithm}}
	\label{fig:figure2}
\end{figure}

To address this issue, DCA leverages **self-supervised learning** to learn from the distribution of the predicted target values. The algorithm dynamically partitions the dataset into smaller, more homogenous subsets and iteratively adjusts the classification boundaries until an optimal segmentation is found. This ensures that each subset contains target values within a narrower range, thereby improving prediction accuracy. The **Dynamic Classification Process** consists of six key modules: **Division Module**, **Classification Model Module**, **Convergence Judgment Module**, **Penalty Module**, **Correction Module**, and **Evaluation Module**, as illustrated in \textbf{Fig.\ref{fig:figure2}}.

During the initialization phase, the training set is divided into $Train_t$ and $Train_p$ using 1:1 ratio, ensuring a balanced training and validation setup.

\begin{figure}[htbp]
	\centering
	\subfigure[N = 5]{\label{fig:figure3_1}\includegraphics[width=0.32\textwidth]{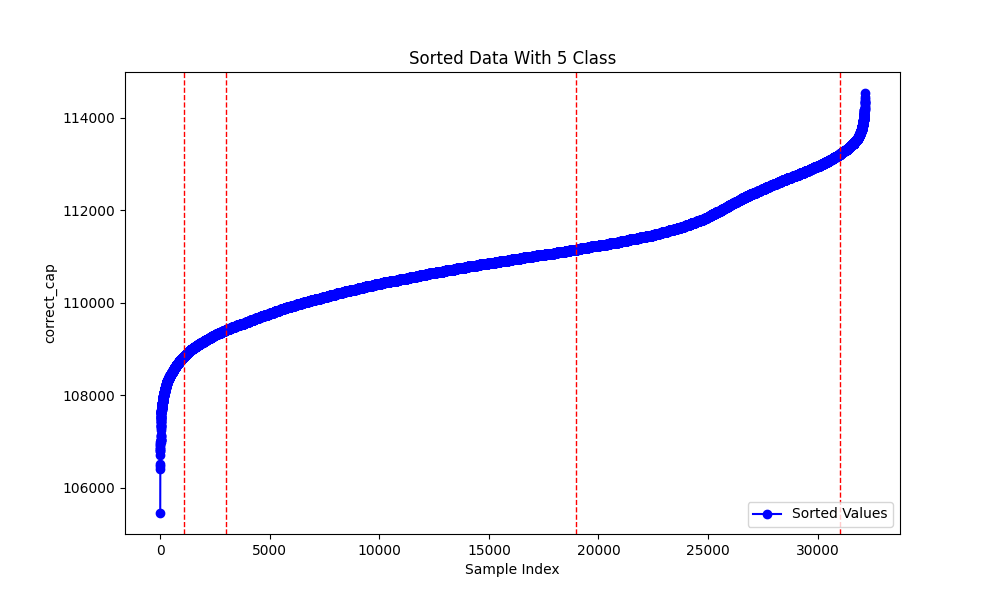}}
	\subfigure[N = 9]{\label{fig:figure3_2}\includegraphics[width=0.32\textwidth]{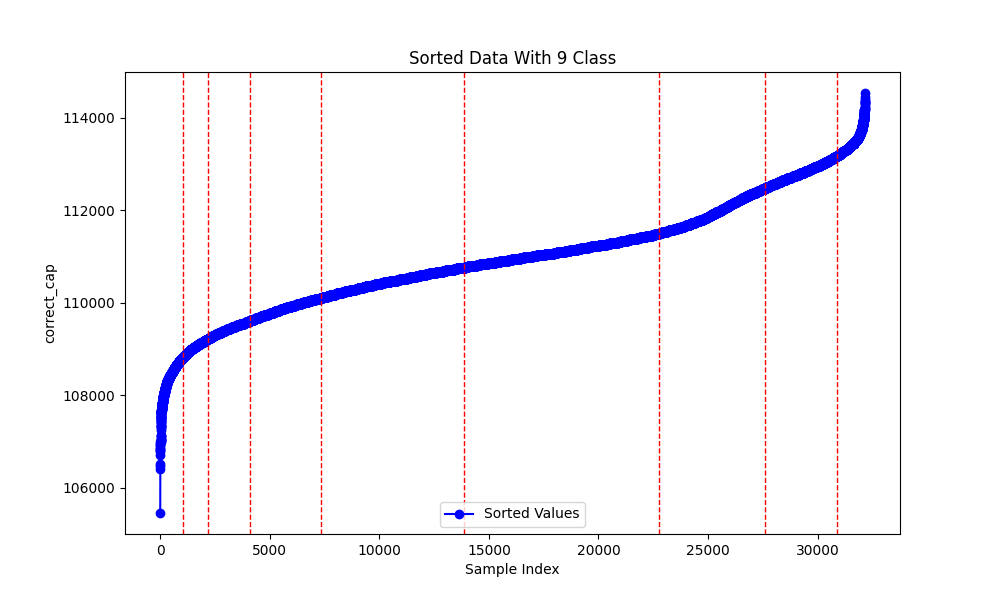}}
	\subfigure[N = 13]{\label{fig:figure3_3}\includegraphics[width=0.32\textwidth]{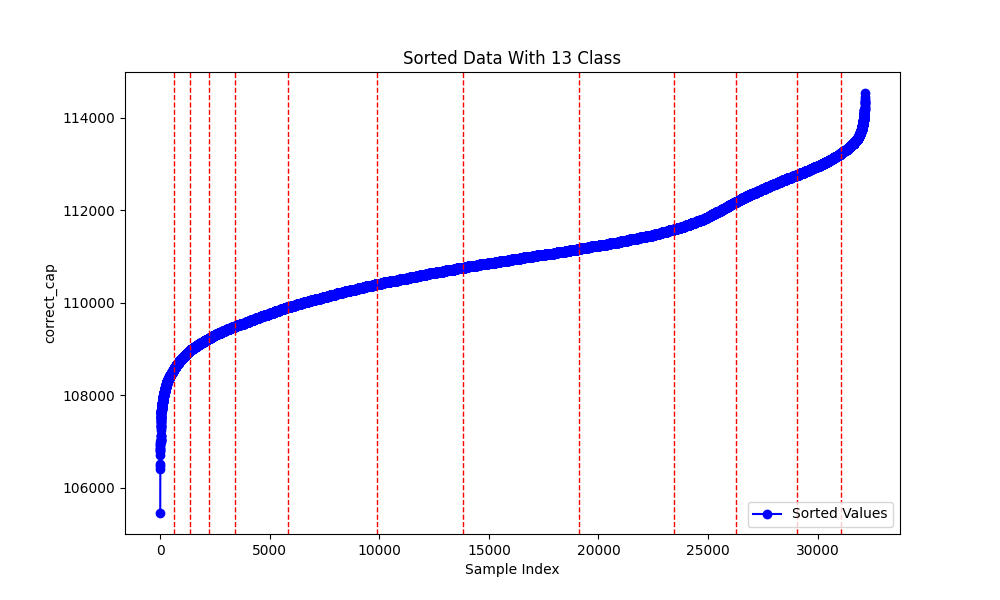}}
	\caption{Classifications $N$ of N = 5, 9, 13}
	\label{fig:figure3}
\end{figure}

\textbf{Division Module}: The first step involves manually defining the number of planned classifications, **N**. The initial classifier applies a **Gaussian kernel function** to measure the distribution of values in the target column of $Train_t$, and the data is then divided based on the degree of fluctuation in the sorted values. Larger fluctuations indicate finer divisions, ensuring more precise segmentation (see \textbf{Fig.\ref{fig:figure3}}). This results in **N intervals**, defined by **N-1 cutoff points**, forming the $Initial Segmentation List$. The **N-1 cutoff points** constitute the initial boundary, and the $Initial Segmentation List$ formed by them will be subsequently employed in the Classification Model Module. The final Boundary will be generated from the $OptimalSegmentationList$, which will be discussed later.

However, in cases where a specific accuracy requirement must be met, manual intervention can yield better results. For example, if only predictions achieving 98\% or higher accuracy are considered valid, the predefined division ranges should be aligned with this requirement. Considering a confidence interval of 95\%, the division module can be manually adjusted such that the preset range aligns with 2\% $\times$ 95\%. This fine-tuning process enhances the precision of the classification boundaries, which in turn optimizes the performance of the \textbf{Exclusion Module}.

\textbf{Classification Model Module}: This module trains a set of classification models using the $Train_t$ and $Train_p$ data. A variety of models—such as **Decision Tree**, **Random Forest**, and **LGBM** are employed, utilizing multi-threading to optimize performance. Each model is trained independently, and its performance evaluated using a confusion matrix on  $Train_p$. The best-performing model is selected for the next iteration based on its ability to minimize classification errors.

\textbf{Convergence Judgment Module}: The confusion matrix output from the Classification Model Module is converted into percentages, with off-diagonal values representing the classification loss. The total loss for each iteration is recorded as **LossList**, and the model with the lowest loss (BestLoss) is identified as the **BestModel**. If the loss values do not change over 15 iterations, or if the last 10 iterations show significant instability, a warning is issued to prevent convergence to a local minimum.

\textbf{Evaluation Module}: To evaluate the model's overall performance, a score is computed using \textbf{Equation~\ref{eq1}}, which balances the best-case loss with the regular performance over the last 10 iterations. This scoring method ensures that models with occasional excellent performance but inconsistent overall results are not selected.
\begin{equation}
	\text{Score} = 0.5 \times \text{BestLoss} + 0.5 \times \frac{\sum_{i=n-10}^{n} a_i}{10}, \quad a_i \in \text{LossList}
	.\label{eq1}
\end{equation}

The $OptimalClassificationModel$ and $OptimalSegmentationList$ are finalized once all iterations are complete.

In the event of non-convergence, the \textbf{Penalty Module }applies penalties based on two core factors: 
\begin{itemize}
	\item Degree of Misclassification: The greater the misclassification degree (i.e., how far the misclassified sample is from its correct classification), the higher the penalty. This is calculated by measuring the distance between the misclassified value and its correct interval, with larger deviations resulting in greater penalties.
	\item Number of Misclassified Samples: The greater the number of misclassified samples in a given class, the stronger the penalty. This is determined by counting the number of samples that do not fit the target classification interval.
\end{itemize}

Partition error stems from features lacking strong support or an unreasonable number of partition intervals. Sometimes, an overly large number of partitions can increase a considerable burden of classification. The penalty is calculated based on the degree of misclassification. Larger deviations from the correct classification resulted in larger penalties. The penalty matrix is updated based Equation \eqref{eq2}:

\begin{equation}Y = \left(\frac{X}{N}\right)^2 + 1.\label{eq2}\end{equation}

where \( X \) represents the degree of misclassification, and \( N \) is the number of classification intervals. This formula ensures that misclassifications with large deviations incur higher penalties.

After calculating the initial penalty, a secondary penalty is applied based on the number of misclassified samples. If a larger number of samples are misclassified in a particular class, the penalty increases. This penalty is calculated using the Equation \eqref{eq2}:
\begin{equation}
	Y = X^2
\end{equation}

This further increases the penalty for large misclassification errors, ensuring that significant discrepancies are addressed.

Once penalties are applied, the misclassified samples are reassessed. If the classification error exceeds an acceptable threshold, the model returns to the \textbf{Classification Model Module} for further iterations and refinement. This iterative process ensures that the model continues to improve until the penalty stabilizes at an acceptable level.

\subsection{Principle of Redundant Prediction}
The best model identified in the Evaluation Module will classify both the training and test sets. The corresponding subsets within their respective intervals are then obtained, as shown in \textbf{Table \ref{tab:classification}}.

\begin{table}[htbp]
	\centering
	\caption{Classification correspondence table}
	\label{tab:classification}
	\scalebox{0.95}{
		\begin{tabular}{cccc}
			\hline
			Set classification & Training set & Prediction model & Prediction set \\
			\hline
			1 & Training set 1 & Model 1 & Prediction set 1 \\
			2 & Training set 2 & Model 2 & Prediction set 2 \\
			\multicolumn{1}{c}{\centering\dots} & \multicolumn{1}{c}{\centering\dots} & \multicolumn{1}{c}{\centering\dots} & \multicolumn{1}{c}{\centering\dots} \\
			N & Training set N & Model N & Prediction set N \\
			\hline
		\end{tabular}
	}
\end{table}

During this process, it is inevitable that there will be slight disturbances to the classification of the test set, meaning that the distribution of the prediction subset and the training subset in the same subset will be slightly different. Generally speaking, the data distribution of the predicted subset is slightly larger than that of the predicted subset, which is caused by fluctuations in the predicted data or minor data input errors. To solve this problem, redundant training was introduced. This technology combines the training data from the target classification interval with the data from adjacent intervals, enabling it to have a more comprehensive training range. Select the data of adjacent intervals based on two criteria:

\begin{itemize}
	\item The data must be sorted by the predicted target values, and the data closest to the target classification is selected.
	\item The amount of data selected from neighboring intervals cannot exceed $K/4$, where $K$ is the total number of target training sets and $4$ is the empirical value obtained from multiple experiments.
\end{itemize}

In this stage, the training subsets within each partition interval will train the same prediction model, such as the linear regression model or the lgmb model, so that each partition interval has the same type but independent model for corresponding predictions. The applying of redundant training enhances the robustness of the model by expanding the data range of the training subset and incorporating relevant data from neighboring intervals. This helps to balance the prediction range of the prediction subset, ensuring that the predicted values within each subset fall within the expected target range, as derived from the $Optimal Segmentation List$ in the Dynamic Classification section. Samples falling outside this range are filtered out, based on the required accuracy, leaving only those predictions that meet the accuracy threshold.

\subsection{Principle of Excluding Predicted Results}
The boundaries of each subset are initially determined by $OptimalSegmentationList$, which is obtained through self-supervised learning from the training set. However, these boundaries can be manually fine-tuned according to the accuracy requirements to achieve a stable performance and higher precision. This ensures that predictions falling within the valid range are retained, whereas those deemed inaccurate are removed.

The exclusion process operates as follows: Once the predictions are generated, they are compared against predefined segmentation boundaries. Predictions that fall outside their corresponding subset ranges are flagged as unreliable and subsequently discarded. This ensures that only the most accurate predictions are retained, thereby enhancing the overall reliability of the classification results.

Critically, the exclusion mechanism is designed to guarantee zero missed detections, meaning that all valid predictions are preserved. Within this constraint, adjustments to segmentation boundaries focus on reducing false positives as much as possible. By dynamically refining these parameters, the exclusion module contributes to a more precise and robust prediction system, allowing for a high accuracy with minimal unnecessary exclusions.

\subsection{Algorithm Advantages and Effectiveness}

In practical applications, prediction-based models face two major challenges. First, it is difficult to balance the impact of samples with lower and higher target values. When the training set is sampled with a normal distribution, the predictions tend to concentrate around the middle of the value range, and using uniform sampling only slightly mitigates this issue. A second challenge arises when attemping to identify inaccurate predictions. After predictions are made, it is necessary to categorize and re-assess them against an accuracy threshold, which often leads to either missed detections or over-exclusion of data.

The DCA effectively addresses these issues by partitioning the data into smaller, well-defined subsets. This segmentation reduces the range within which each model operates, allowing it to focus on narrower and more homogeneous data intervals. Consequently, the complexity of the predictions is minimized,thereby improving the accuracy of the model. Importantly, this approach ensures that each subset operates independently, reducing errors that might arise from overlapping data distributions or perturbations in the predictions.

Without dynamic classification, unsupervised models may fail to properly distinguish between features with similar values, leading to inaccurate predictions, particularly when feature distributions are broad. Moreover, in traditional prediction structures, the re-learning and selection of new models after predictions often exacerbate these issues, making them computationally expensive and inefficient.

The DCA effectively reduces the complexity of predictions by limiting the data range of each subset. By segmenting the data into quantifiable intervals, DCA can precisely filter the prediction results, ensuring that only predictions that meet the required accuracy are retained. This method is particularly significant in industrial applications, because it eliminates the need for additional models or complex retraining steps, thereby simplifying the prediction process. 

Furthermore, DCA ensures that the model achieves the goal of zero missed detections while minimizing false positives as much as possible. By carefully adjusting the segmentation boundaries, the algorithm allows for stable and highly, accurate predictions with minimal exclusion. This balancing act is crucial in real-world scenarios, where the precision of the classification process is paramount, and the model must maintain both accuracy and robustness under various operational conditions.

In practical terms, this algorithm has demonstrated its effectiveness by achieving a high accuracy with minimal exclusion. Through the use of self-supervised learning and optimal segmentation, DCA reliably filters predictions, achieving zero missed detections while minimizing false positives. This ability to discard substandard predictions directly based on data characteristics ensures that the model remains robust and accurate, even in real-world, dynamic environments.

The DCA has shown superior performance on various datasets, offering a promising solution for applications that require high precision, such as industrial forecasting. Although the algorithm still presents opportunities for improvement, particularly in automatic parameter tuning and classification efficiency, it has already proven to be a valuable tool for improving prediction accuracy in both small and large-scale classification tasks.

\section{Data Testing Situation}
\subsection{Non-Open Source Data}

Consistency is crucial in battery production. In production processes involving the chemical and sorting stages, cells must undergo discharge testing to verify that their capacity meets the required standards. The test data for this study come from actual factory production, consisting of three different production batches, and is used to predict the cell capacity. The necessary information about these batches has been anonymized and is provided in \textbf{Table \ref{tab:batch_data_info}}.

\begin{table}[h]
	\centering
	\caption{Non-Open Source Data Information}
	\scalebox{0.95}{
		\begin{tabular}{|c|c|c|c|}
			\hline
			Batch/Dataset & 22LA & 27LA & 13LA \\ \hline
			Production Time & May (15 days) & June (26 days) & July (6 days) \\ \hline
			System Information & System 1 & System 2 & System 3 \\ \hline
			Approximately Data & 70,000 & 140,000 & 85,000 \\ \hline
		\end{tabular}
	}
	\label{tab:batch_data_info}
\end{table}

The data for these three batches follow a normal distribution. Feature extraction primarily relies on characteristics from the middle production stages, such as the weight, time, density, and voltage features from the formation and capacity sorting stages. Data preprocessing steps include filtering, temperature compensation, and outlier removal using the interquartile range method. The dynamic classification algorithm can be applied to both uniform and normal distributions, although the parameter adjustments may vary.

\subsubsection{Comparison Plan and Indicator Settings}

Owing to confidentiality, only the specific numerical values can be approximated. The total number of features is approximately 100, of which approximately 20\% originate from intrinsic battery cell characteristics (such as weight and coating density), whereas the remaining 80\% are derived from data generated during the production process, including voltage, temperature, and capacity data. In the data preprocessing stage, outliers are handled using the interquartile range method. Specifically, 17\% of the outliers are filtered from the 22LA, 19\% from the 27LA dataset, and 14\% from the 13LA dataset respectively. The remaining are normalized and standardized. Missing values are removed from the dataset, and no duplicates are found. Although other outlier detection methods are considered, the interquartile range (IQR) method is deemed sufficient for this experiment owing to its robustness. The IQR method is particularly effective in handling skewed distributions and is less sensitive to extreme values, making it a suitable choice for ensuring the stability and accuracy of the model in this case. This method effectively isolates the core data distribution, ensuring that only significant deviations are treated as outliers.

Because the average accuracy alone is not sufficient to effectively differentiate between models, this study selects more detailed indicators to assess model performance. These include the proportion of samples with prediction accuracy between 99\% and 100\% relative to the total, and the proportion of samples with a prediction accuracy between 99.5\% and 100\%. The higher the percentage of these two indicators, the greater the reliability and precision of the prediction.

The experiment compared three methods: K-means model + prediction, GMM model + prediction, and the dynamic classification model. All of these models use linear regression as the underlying prediction method, while the direct prediction result (DP) of the linear model is also recorded as comparison data. The training and test sets are divided in a 3:7 ratio, using the same features and outlier detection methods for both sets. To ensure reliability, each experiment is verified using two random seeds, generating two sets of random test results.

For subsequent tests, all prediction models use the same features and parameters, with linear regression serving as the comparison method to further assess performance differences. These detailed indicators allow for a more accurate evaluation of each model performance.

\subsubsection{Comparison Analysis}
\textbf{Table \ref{tab:best_prediction_results}} records all test results with the best prediction effect.
\begin{table}[htbp]
	\centering
	\caption{Best Prediction Results for Non-Open Source Data}
	\label{tab:best_prediction_results}
	\scalebox{0.82}{\begin{tabular}{cccccccccc}
			\hline
			\textbf{Dataset} & \textbf{Item} & & \textbf{DP} & & \textbf{GC} & & \textbf{KC} & & \textbf{DC}\\			
			\hline
			\multirow{2}{*}{22LA} & 1\% Error Ratio & & 96.69\% & & 98.40\% & & 97.68\% & &99.56\%\\
			
			& 0.5\% Error Ratio & & 78.94\% & &85.7\% & &82.89\% & &98.15\% \\
			\hline
			\multirow{2}{*}{27LA} & 1\% Error Ratio & &92.52\% & &94.37\% & &94.03\% & &99.56\% \\
			& 0.5\% Error Ratio & &65.87\% & &70.98\% & &70.58\% & &96.73\% \\
			\hline
			\multirow{2}{*}{13LA} & 1\% Error Ratio & &97.72\% & &98.03\% & &97.79\% & &99.75\% \\
			& 0.5\% Error Ratio & &86.37\% & &87.19\% & &86.29\% & &98.94\%\\
			\hline
	\end{tabular}}
\end{table}

After completing these steps, to achieve the goal of zero missed detections and minimal false positives, we must exclude inaccurate predictions. To evaluate this step, we introduce comparisons with advanced models such as XGBoost, Gaussian classification models, and Random Forest. As advanced methods, these models are more capable of capturing detailed information, which typically results in better performance.

\begin{figure}[h!]
	\centering
	\subfigure[Excluding results of XGBoost]{\label{fig:figure4_1}\includegraphics[width=0.95\textwidth]{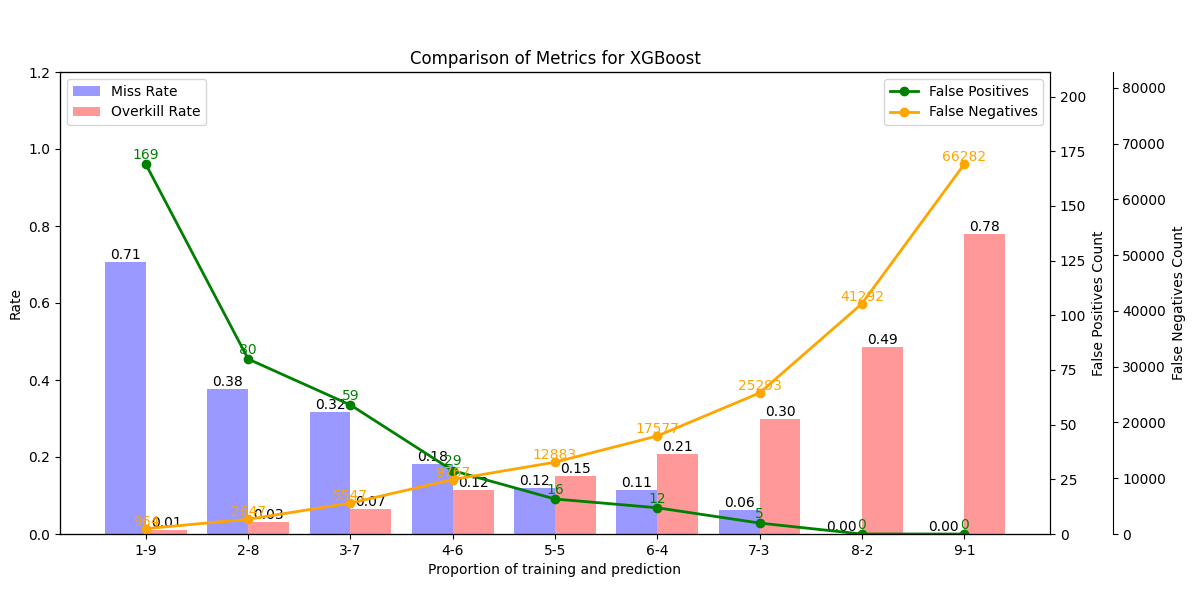}}
	\subfigure[Excluding results of Gaussian SVM]{\label{fig:figure4_2}\includegraphics[width=0.475\textwidth]{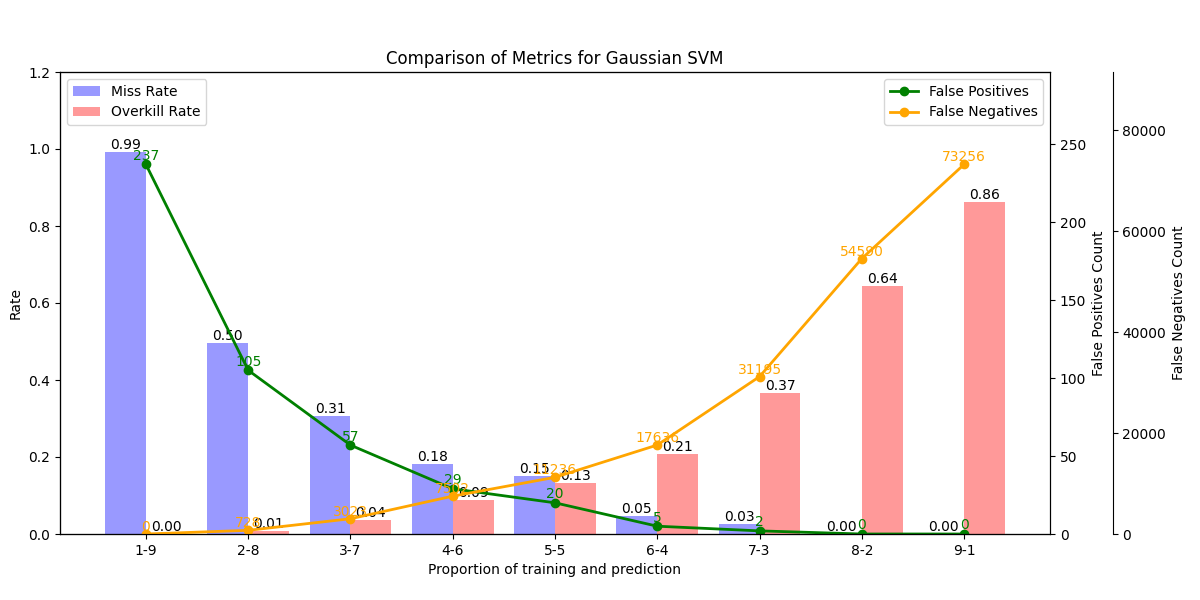}}
	\subfigure[Excluding results of Random Forest]{\label{fig:figure4_3}\includegraphics[width=0.475\textwidth]{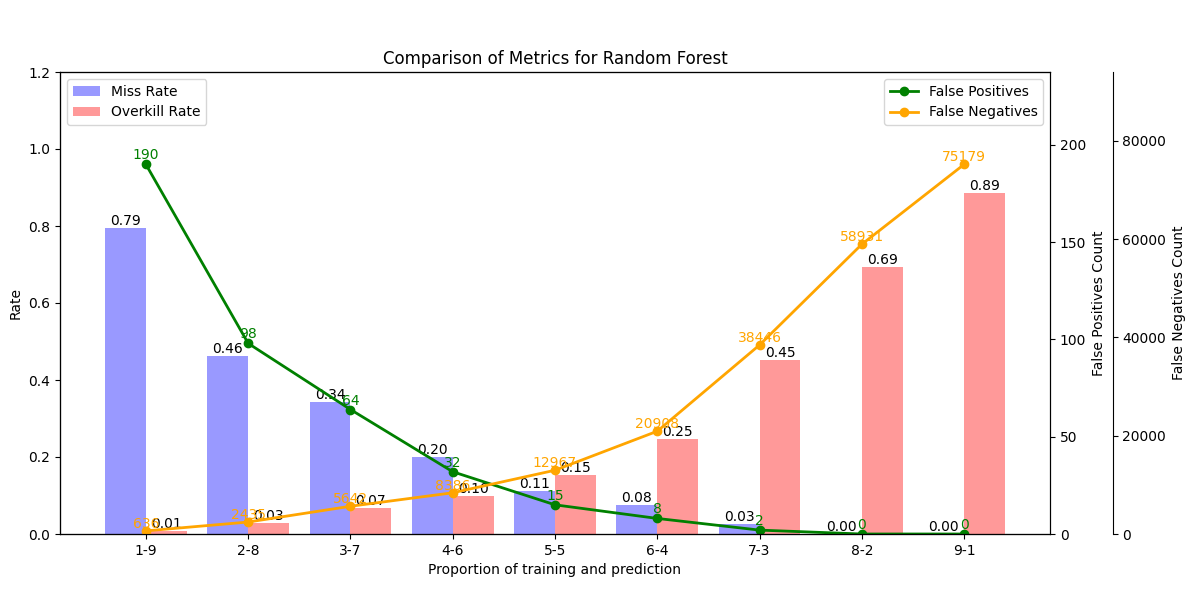}}
	\caption{Excluding Results of Three Models}
	\label{fig:figure4}
\end{figure}

Because introducing a new model requires a certain amount of data for training, let us assume that some data are predicted inaccurately, and this data will serve as the training set. This assumption makes to better illustrate the advantages of the dynamic classification algorithm in its mechanism. Assume that the data with prediction accuracy below 99\% are $J$, and we divide $J$ into 10 parts, supplementing them with randomly select data that meet the accuracy threshold, ensuring that the training set size remains constant at $J$, with consistent features. Each model uses 9 different training sets, where the ratio of inaccurate to accurate data starts at 1:9 and gradually increases until it reached 9:1. Finally, the missed detection rate, false positive rate, missed detection count, and false positives count are recorded. The results of the three models are shown in \textbf{Fig.\ref{fig:figure4}}.

Among the experimental results from these three models, all models achieve a good balance with a 4:6 training-to-test data ratio. The corresponding metrics for false positives and false negatives are shown in the graph. The bar sections represent the miss rate (blue) and overkill rate (red), whereas the line sections indicate the absolute number of false positives and false negatives. We can observe that as the proportion of non-compliant data in the training set increases, the miss rate significantly decreases, but the overkill rate gradually increases. This result is the combined effect of changes in both the proportion of the training set and non-compliant data in the test set.

In terms of achieving the goal of 'zero missed detections and minimal false positives', XGBoost performed the best. When the proportion of non-compliant to compliant data was set at 8:2, XGBoost successfully excluded all non-compliant data, as indicated by the low miss rate. However, the false positive rate remained significant, and the number of misclassified samples was also notable. These results illustrate the trade-off in balancing the two metrics: although misclassified instances were minimized, almost half of the remaining data were misidentified. This discrepancy highlights that, in real-world scenarios, it is difficult for the model to fully achieve the ideal of zero missed detections and minimal false positives, as production data often contain noise and errors that are difficult to discern. The graph visually represents the performance of the model under different training-prediction ratios, the progression from a high miss rate to an acceptable level as the proportion of non-compliant data decreases.

Under the DCA framework, the model leverages information derived from effective self-supervised learning to filter out predictions with significant errors. After the independent prediction, each predicted result falls within its designated range. If a sample exceeds this range by more than ±5\%, it is considered inaccurate. If there are specific accuracy requirements, the range can be fine-tuned by adjusting the segmentation values, thus achieving zero missed detections and minimal false positives. This method ensures that the boundaries of each subset are confined within accuracy limits, ensuring that each prediction meets the desired precision standard.

The range for each prediction is derived from OptimalSegmentationList, which is initially defined by the initial segmentation values. While the division module automatically generates initial segmentation values based on the distribution of the training set samples, manual adjustments of these initial values can provide results that better meet specific accuracy requirements.

\begin{figure}[htbp]
	\centering
	\subfigure[13LA Non-excluded ]{\label{fig:figure5_1}\includegraphics[width=0.475\textwidth]{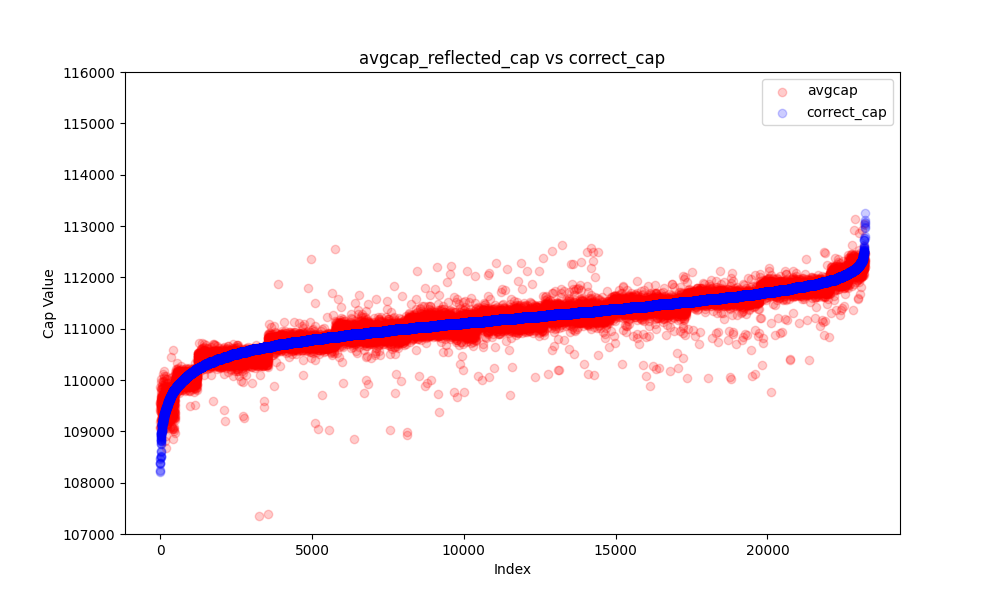}}
	\subfigure[13LA Excluded]{\label{fig:figure5_2}\includegraphics[width=0.475\textwidth]{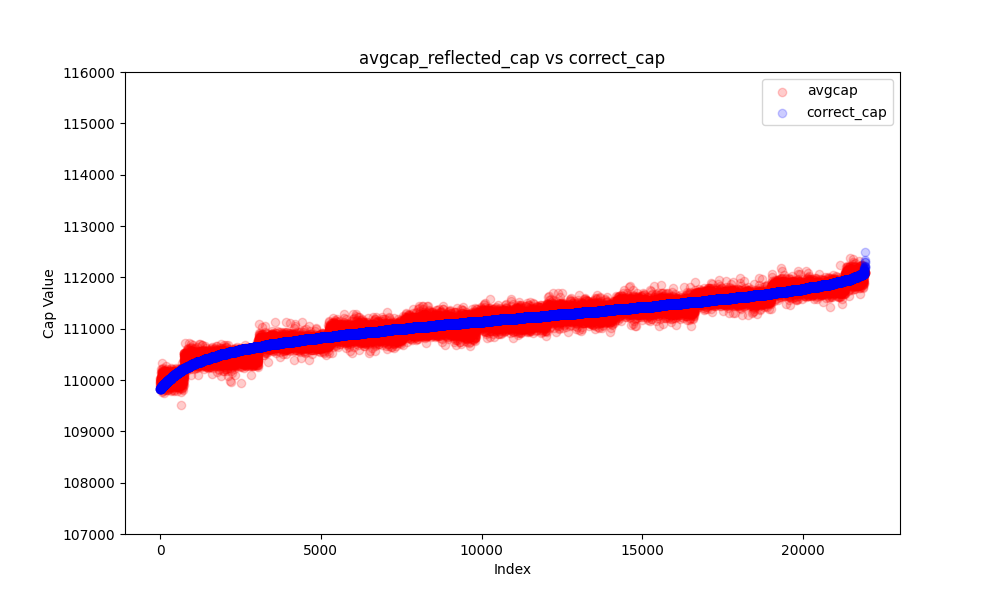}}
	\subfigure[13LA Comparison]{\label{fig:figure5_3}\includegraphics[width=0.95\textwidth]{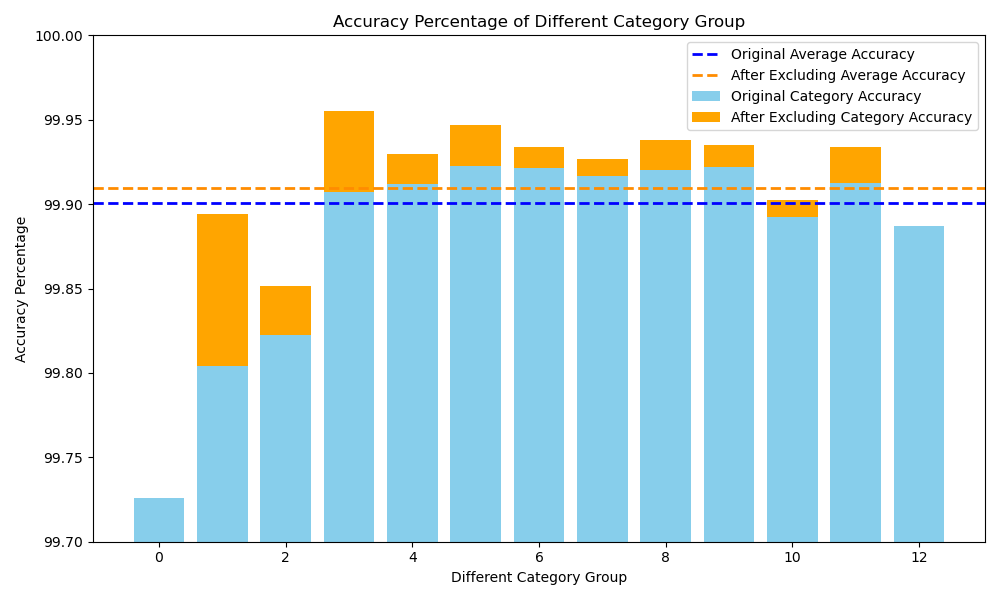}}
	\caption{Average precision results}
	\label{fig:figure5}
\end{figure}

In the 13LA dataset, to directly filter the data with an error of less than 1\%, the predictions were excluded using the exclusion module of the dynamic classification to obtain the final prediction. First, the initial division ratio of dynamic classification was manually set as:
[0.02, 0.03, 0.1, 0.1, 0.1, 0.1, 0.1, 0.1, 0.1, 0.1, 0.1, 0.1, 0.03, 0.02]
Finally, the LGBT classification model achieved the best result, and the optimized division ratio, combined with the numerical distribution of the targets in the training set, produced the OptimalSegmentationList:[108873.9, 109546.5, 110193.2, 110485.0, 110707.1, 110873.8, 111040.5, 111193.3, 111359.85, 111540.5, 111860.0, 112123.8]
The classification error was only 3.39\%. Prediction results were collected and matched to the division list for exclusion. The exclusion was set to entirely discard the first and last categorization intervals, while the 11 intermediate intervals were expanded by 1.0025 times each. The non-excluded and excluded results are shown in \textbf{Table \ref{tab:Culled Result}}, and the number of samples per interval is shown in \textbf{Table \ref{tab:Culled Information}}. The obtained image results are shown in \textbf{Fig.\ref{fig:figure5}}

\begin{table*}[htbp]
	\centering
	\caption{Samples Number Table of Non-excluded and Excluded Information for 13LA}
	\label{tab:Culled Information}
	\scalebox{0.5}{
		\begin{tabular}{|c|c|c|c|c|c|c|c|c|c|c|c|c|c|c|c|}
			\hline
			\textbf{Item} & \multicolumn{13}{c|}{\textbf{Number of samples per interval}} & \textbf{Total number}& \textbf{Proportions}\\
			\hline
			Non-excluded Result &818&2750&8293&8200&8918&8287&8618&9853&9141&9027&7991&2329&1002&85227&100.0000\%\\
			\hline
			Excluded Result &0&2708&8170&8023&8677&8077&8383&9585&8934&8875&7863&2288&0&81583&95.7244\%\\
			\hline
			Excluded Samples Number &818&42&123&177&241&210&235&268&207&152&128&41&1002&3644&4.2756\%\\
			\hline
		\end{tabular}
	}
\end{table*}

\begin{table}[htbp]
	\centering
	\caption{Comparison Table of Non-excluded and Excluded Information for 13LA}
	\label{tab:Culled Result}
	\scalebox{0.9}{
		\begin{tabular}{|>{\centering\arraybackslash}p{80pt}|>{\centering\arraybackslash}p{70pt}|>{\centering\arraybackslash}p{90pt}|>{\centering\arraybackslash}p{95pt}|}
			\hline
			\textbf{Item} & \textbf{Average Accuracy} & \textbf{Percentage of errors within 1\%} & \textbf{Percentage of errors within 0.5\%} \\
			\hline
			Non-excluded Result & 99.9142\% & 99.9879\% & 99.7495\% \\
			\hline
			Excluded Result & 99.9148\% (0.0006\%↑) & 100.0000\% (0.0121\%↑)& 99.7941\% (0.0446\%↑) \\		
			\hline
		\end{tabular}
	}
\end{table}

Because the error of the dynamic classification is very small, this method allows us to discard 5\% of the data and still achieve 99\% or higher accuracy for all the remaining data.

From the above results, it is evident that the DCA has several advantages:

\begin{itemize}
	\item High Prediction Accuracy: DCA significantly improves prediction accuracy by dividing the data into smaller subsets and making predictions within smaller, controlled ranges, thereby reducing prediction errors.
	\item Effective Exclusion Mechanism: The exclusion mechanism ensures that predictions outside the expected range are filtered out, achieving zero missed detections and minimal false positives, as demonstrated in the 13LA dataset.
	\item No Need for Additional Models: Unlike traditional methods, DCA does not require additional models or retraining steps, reducing computational overhead while maintaining high prediction accuracy.
	\item Flexible and Adaptable: DCA's ability to dynamically adjust the classification ranges and fine-tune the model makes it highly adaptable to various datasets and prediction tasks, ensuring robust performance even in complex scenarios.
\end{itemize}

\begin{table*}[htbp]
	\centering
	\caption{Prediction Results for Open Source Data}
	\label{tab:open_source_data_results}
	\scalebox{0.45}{
		\begin{tabular}{|c|ccccccccc|ccc|}
			\hline
			\textbf{Dataset} & \textbf{Item} & \textbf{RF} & \textbf{SVM} & \textbf{Linear} & \textbf{LGBM} & \textbf{BP} & \textbf{XGBoost} & \textbf{DC} & \textbf{DC-E} & \textbf{N} & \textbf{DC Error} &\textbf{Excluded Rate}\\
			\hline
			\multirow{2}{*}{California Housing Prices} & MSE & 0.0088 & 0.0181 & 0.0219 & 0.0109 & 0.0137 & 0.0087 & 0.0102 & 0.0093 & \multirow{2}{*}{N=3} & \multirow{2}{*}{14.96\%} & \multirow{2}{*}{5.31\%} \\
			& R2 & 0.7756 & 0.5377 & 0.6149 & 0.7210 & 0.6514 & 0.7772 & 0.7411 & 0.7448& & &\\
			\hline
			\multirow{2}{*}{Appliances Energy Prediction} & MSE & 0.0043 & 0.0097 & 0.0092 & 0.0058 & 0.0055 & 0.0046 & 0.0052 & 0.0042 & 
			\multirow{2}{*}{N=4} & \multirow{2}{*}{34.59\%} & \multirow{2}{*}{ 28.11\%} \\
			& R2 & 0.6003 & 0.1103 & 0.1569 & 0.4651 & 0.4894 & 0.5793 & 0.5244 & 0.5771 & & &  \\
			\hline
			\multirow{2}{*}{Estimation Obesity} & MSE & 0.0001 & 0.0036 & 0.0061 & 0.0001 & 0.0001 & 0.0001 & 0.0002 & 0.0001 & 
			\multirow{2}{*}{N=4} & \multirow{2}{*}{1.40\%} & \multirow{2}{*}{16.20\%} \\
			& R2 & 0.9936 & 0.8036 & 0.6610 & 0.9950 & 0.9960 & 0.9932 & 0.9917 & 0.9961 & & &  \\
			\hline
			\multirow{2}{*}{Student Performance Prediction} & MSE & 0.0094 & 0.0205 & 0.0141 & 0.0095 & 0.0253 & 0.0097 & 0.0125 & 0.0114 & 
			\multirow{2}{*}{N=7} & \multirow{2}{*}{45.93\%} & \multirow{2}{*}{27.68\%} \\
			& R2 & 0.8555 & 0.6865 & 0.7842 & 0.8544 & 0.6128 & 0.8508 & 0.8085 & 0.8488 & & & \\
			\hline
		\end{tabular}
	}
\end{table*}

It is important to note that for DCA to achieve such excellent results, the classification errors must remain below approxiamtely 5\%. If classification errors exceed this threshold, unexpected samples may appear in the classification results, distorting the range of classification intervals. This disruption in the classification range would lead to a decline in the overall prediction accuracy making it difficult to achieve the goal of zero missed detections and minimal false positives. Therefore, maintaining the classification error below 5\% is crucial for the performance of the algorithm.

\subsection{Open Source Data}
The open-source datasets are sourced entirely from Kaggle. Four non-sequential datasets, representing various fields, are selected to validate the performance of the proposed model. Comparisons are made with the Random Forest model, SVM (with a Gaussian kernel), Linear Regression model, LGBM model, BP Neural Network model, XGBoost model, and the DC algorithm.Four datasets from different domains were chosen for the regression prediction task:
\begin{enumerate}
	
	\item California Housing Prices dataset: a widely used dataset for regression tasks. It contains over 20,000 entries but suffers from a data availability rate of only 85.3\%. This dataset is well-suited for verifying the prediction accuracy of the models.
	
	\item Appliances Energy Prediction: This dataset contains data related to energy usage and environmental conditions within a residential building. It includes measurements of the temperature, humidity, and energy consumption of appliances and lights in various rooms, as well as outdoor weather data from a nearby weather station.
	
	\item Estimation of Obesity: This dataset includes data for estimating obesity levels in individuals from Mexico, Peru, and Colombia based on their eating habits and physical condition. It contains both real and synthetically generated data.
	
	\item Student Performance Dataset: This dataset contains information on student performance in Math and Portuguese language courses, including demographic, family, and school-related features.
\end{enumerate}

\begin{figure}[htbp]
	\centering
	\subfigure[California Housing Prices]{\label{fig:figure8_1}\includegraphics[width=0.23\textwidth]{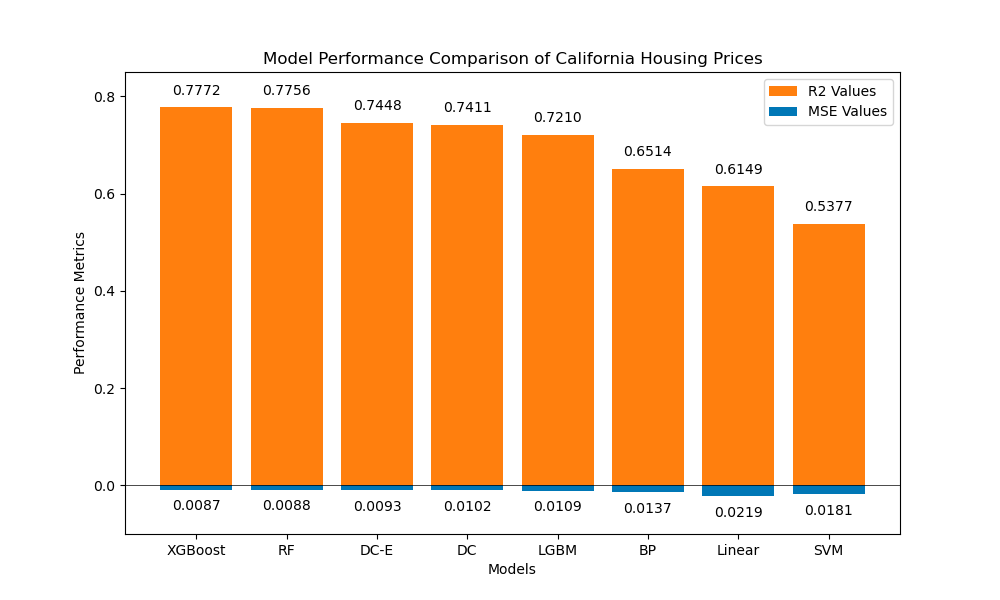}}
	\subfigure[predicting energy usage]{\label{fig:figure8_2}\includegraphics[width=0.23\textwidth]{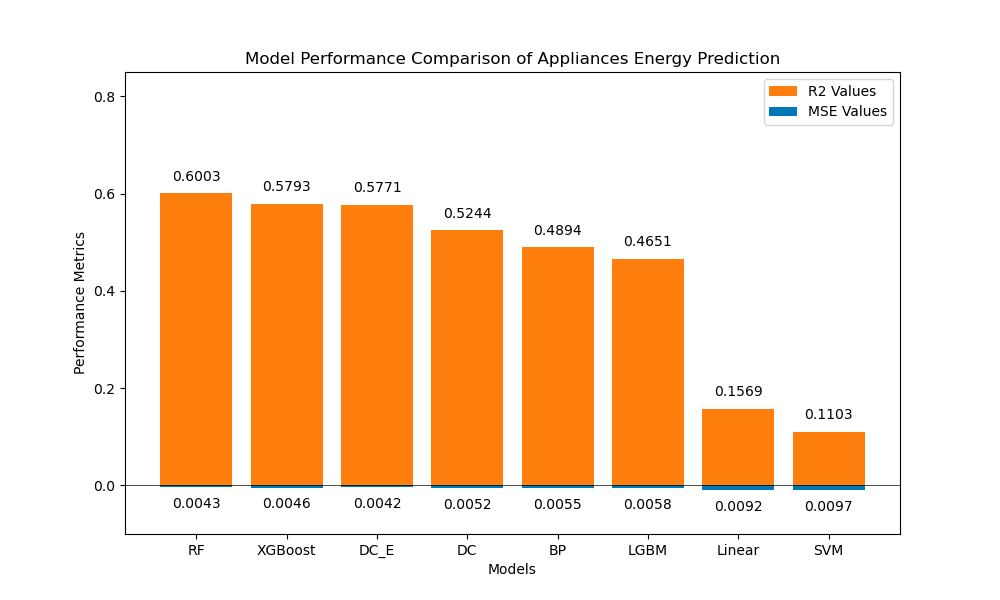}}
	\subfigure[Estimation Obesity]{\label{fig:figure8_3}\includegraphics[width=0.23\textwidth]{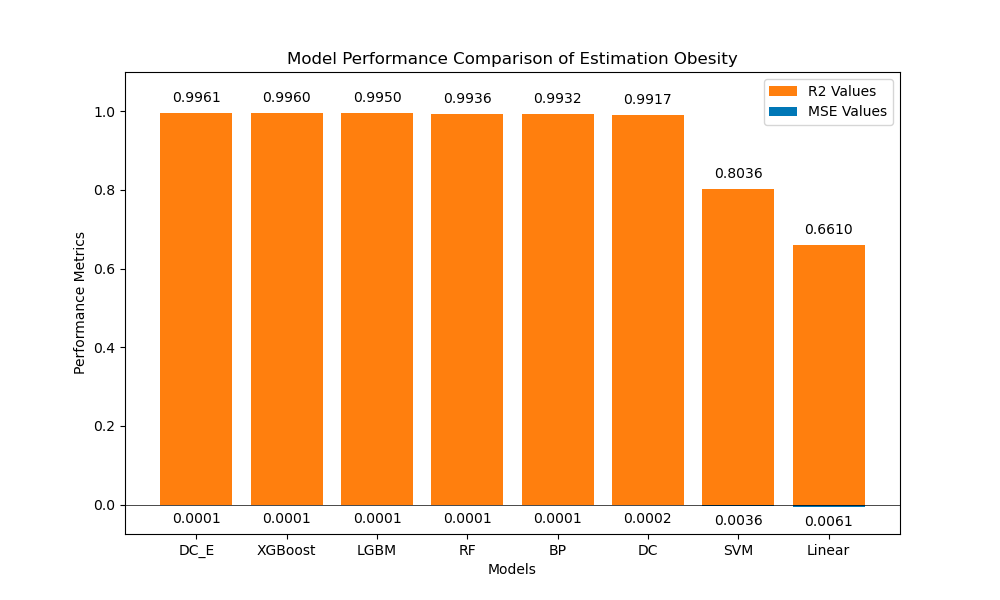}}
	\subfigure[predicting student scores]{\label{fig:figure8_4}\includegraphics[width=0.23\textwidth]{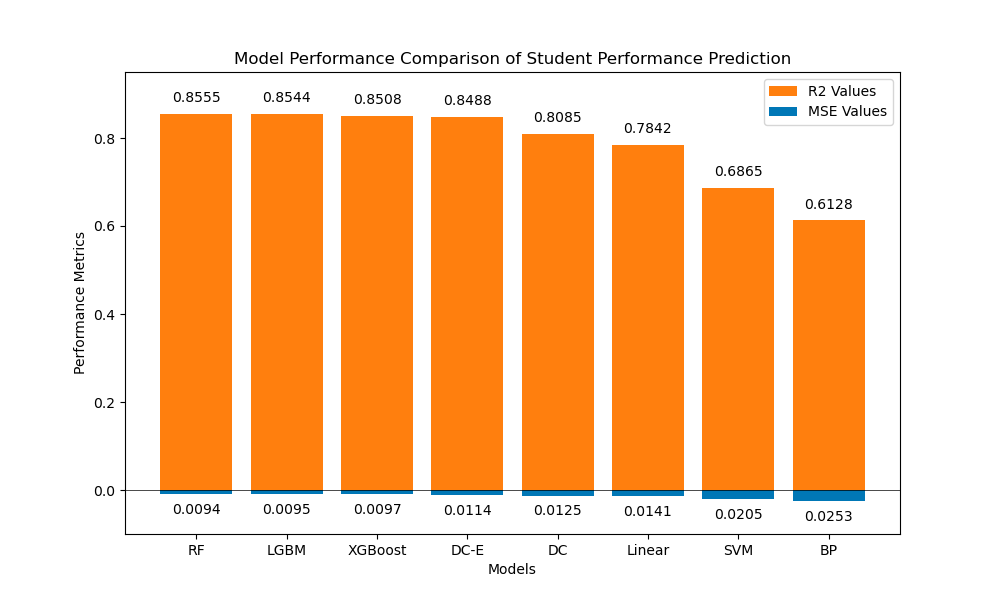}}
	\caption{Prediction results under two different public datasets}
	\label{fig:figure6}
\end{figure}

Due to the relatively clean data, these datasets were processed within a uniform framework, employing outlier detection through a quartile-based approach and normalization. The training and test sets were divided in a 1:1 ratio. The details of the data, including the effects of classification errors and culling features, are presented in \textbf{Table \ref{tab:open_source_data_results}} and \textbf{Fig.\ref{fig:figure6}} shows the evaluation results.

In our previous research, we demonstrated that introducing a new model to achieve zero false negatives and minimal false positives was an unwise choice, both in terms of model performance and training set requirements, which led us further from our objective. Closed-source data typically originates from manufacturers with stable production processes and strong management, which results in higher data purity and consistency, generally yielding higher-quality data. Such data are often well-standardized and consistent, making them more suitable for direct analysis and modeling, typically leading to reliable outcomes. In contrast, open-source datasets are generally associated with greater uncertainty because they may originate from various production or testing environments, with the quality of data potentially varying significantly between sources. Consequently, open-source datasets tend to be more complex and challenging. Therefore, this part of the research focuses on analyzing the impact of feature exclusion versus inclusion using the DCA model on prediction performance.

The DCA performed admirably across all four datasets, achieving top-tier results, particularly when the classification error was below 5\%. However, because the DC algorithm relies on a classification method for subclass prediction, its performance is impacted when the feature correlation is weak, leading to higher classification errors. In such cases, even after some data are excluded, the results only approach the performance of Random Forest and XGBoost. When analyzing the Estimation Obesity dataset, which benefits from stronger feature support, the classification error was very low. This indicates that the samples in each subset are nearly always accurate, thereby allowing the $ OptimalSegmentationList$ to effectively support the culling function. This enables data reorganization without requiring additional models, thereby significantly improving prediction results. Conversely, datasets such as appliances energy prediction and student performance prediction, which have higher classification errors, lack strong feature support, making it difficult to surpass the performance of Random Forest and XGBoost. Nonetheless, the DCA shows great potential, particularly when classification errors are low, allowing it to filter out low-accuracy data through hyperparameter design and the culling function, thereby enhancing the final prediction results.

\section{Summary and Next Steps}
This study introduces the DCA that designed to address the limitations of existing prediction models. By leveraging self-supervised learning based on the distribution of sample data in the training set, the algorithm effectively narrows the range of data within each subset, leading to improved prediction accuracy. The algorithm consists of three key components: regularity, classification, prediction. Each categorized subclass, containing both predicted and training data, allows the algorithm to independently predict smaller ranges, thereby enhancing the overall prediction results. Consequently, DCA is highly sensitive to classification errors, and when these errors are minimized, the algorithm demonstrates an excellent performance.
In the case of non-public datasets, particularly factory production data, DCA achieved a classification error rate of less than 1\%, significantly enhancing the prediction accuracy. For public datasets, although the classification errors were slightly higher, the performance of the DCA remained comparable to that of other advanced models, including Random Forest and XGBoost. Notably, in the Estimation Obesity dataset, where the classification error rate was only 1.4\%, DCA delivered exceptional results, confirming its ability to utilize the self-supervised information of DCA to effectively support data filtering. In turn, this allowed the algorithm to outperform existing solutions.

Despite its promising results, the DCA algorithm still requires optimization, particularly in the classification phase. Future resea rch will focus on these three primary optimization areas:

\begin{enumerate}
	\item Automatic Parameter Adjustment: Develop an automated system capable of dynamically adjusting the number and sensitivity of classifications according to the data distribution. This adjustment aims to achieve more accurate segmentation and improve the classification performance. In addition, explore the possibility of designing a mechanism that enables the DCA to flexibly select the prediction model for each subset within the Redundant Prediction section. The integrated hyperparameter framework is also theoretically feasible. Currently, a fixed prediction model is manually selected, which may limit the prediction accuracy. By enabling flexible model selection, better prediction results can potentially be achieved.
	\item Classification Model Dependency: Enhancing the algorithm's robustness by reducing its dependence on the effectiveness of classification model. A stronger classification model will leads to better results, particularly when the feature correlation is weak.
	\item Avoiding Local Optima: Introducing mechanisms to prevent the algorithm from falling into local optima and ensuring that the model converges to an optimal solution.
\end{enumerate}
These optimizations will further improve the robustness and applicability of the DCA, enabling it to handle a wider range of data characteristics and real-world scenarios, and provide even more reliable and accurate predictions.

\end{document}